\documentclass[letterpaper, 10 pt, conference]{ieeeconf}  % Comment this line out if you need a4paper

\IEEEoverridecommandlockouts                              % This command is only needed if 
\usepackage{graphics} % for pdf, bitmapped graphics files
\usepackage{epsfig} % for postscript graphics files
\usepackage{mathptmx} % assumes new font selection scheme installed
\usepackage{times} % assumes new font selection scheme installed
\usepackage{amsmath} % assumes amsmath package installed
\usepackage{amssymb}  % assumes amsmath package installed
\usepackage{booktabs}
\usepackage{float}
\usepackage{caption}
\usepackage{cite}
\usepackage[colorlinks,bookmarksopen,bookmarksnumbered,citecolor=blue,linkcolor=blue,urlcolor=blue]{hyperref}
\title{\LARGE \bf
Lightweight Safe Reinforcement Learning for End-to-End UAV Navigation
}

\author{
Shenghui Zhang,
YuXuan Gao,
Songwei Zhao,
Jifeng Hu,
Zijing Zhang,
and Hechang Chen
}
\begin{document}

\maketitle
\thispagestyle{empty}
\pagestyle{empty}

%%%%%%%%%%%%%%%%%%%%%%%%%%%%%%%%%%%%%%%%%%%%%%%%%%%%%%%%%%%%%%%%%%%%%%%%%%%%%%%%
\begin{abstract}

With the rapid development of autonomous aerial systems, Unmanned Aerial Vehicles (UAVs) are increasingly deployed in applications such as inspection, environmental monitoring, and rescue, creating growing demand for reliable autonomous navigation. However, autonomous UAV navigation in dense environments remains challenging under sparse perception and dynamic constraints. Most reinforcement learning (RL) methods lack explicit safety mechanisms, leading to unsafe exploration, unstable training, and risky behaviors, especially during high-speed flight. Even in safe RL approaches, safety is often enforced by projecting policy outputs onto a safe action set, which may introduce instability. Meanwhile, many learning-based methods rely on dense inputs or large networks, increasing computational burden and limiting lightweight onboard deployment. Facing the above challenges, we propose a safety-constrained perception-control integrated framework for UAV navigation. A lightweight network encodes sparse observations into collision-risk-aware features using asymmetric and depthwise separable convolutions. We formulate the task as a constrained Markov decision process within a hierarchical control architecture and solve it using a Lagrangian-based safe PPO algorithm. Curriculum learning further improves training stability. Experiments with varying obstacle densities and flight speeds demonstrate higher success rates, improved safety, and better efficiency than existing reinforcement learning baselines.

\end{abstract}

%%%%%%%%%%%%%%%%%%%%%%%%%%%%%%%%%%%%%%%%%%%%%%%%%%%%%%%%%%%%%%%%%%%%%%%%%%%%%%%%
\section{INTRODUCTION}

With the rapid development of avionics, sensing technologies, and intelligent control, Unmanned Aerial Vehicles (UAVs) have been widely deployed in applications, such as aerial inspection, logistics delivery, and emergency response\cite{mao2025robust},\cite{guo2023uav}. In recent years, the rapid growth of the low-altitude economy has further expanded UAV application scenarios, which also places higher requirements on autonomous flight and safe navigation capabilities in complex environments.

Recent studies on UAV autonomous navigation can be broadly categorized into model-based planning methods and learning-based approaches. Classical planning algorithms, such as A*, RRT, and their variants, generate collision-free trajectories using known or reconstructed environment maps and have demonstrated strong performance in structured environments\cite{mao2025robust},\cite{cao2025modified},\cite{zammit2022comparison}. However, in complex and unstructured scenarios with dense obstacles and incomplete perception, these methods often suffer from limited robustness and high computational cost due to frequent replanning. 

Reactive control and learning-based approaches have been introduced to address these limitations.In particular, reinforcement learning (RL) enables UAV to learn navigation policies directly through environment interaction and has shown promising results using algorithms such as DQN\cite{guo2023uav}, DDPG\cite{bouhamed2020autonomous}, TD3\cite{zhang2022autonomous}, SAC\cite{lei2025drl}, and PPO\cite{tai2017virtual}. Meanwhile, LiDAR-based perception has been widely adopted due to its robustness to illumination changes and accurate geometric sensing\cite{cao2025modified},\cite{lakshmi2024enhancing}. Several studies leverage raw range measurements or transformed representations (e.g., polar maps or depth images) as inputs to policy networks, enabling end-to-end obstacle avoidance\cite{lei2025drl},\cite{tai2017virtual}.

Despite these advances, several challenges remain unsolved, especially under sparse perception and high-speed navigation conditions. First, existing approaches rely on dense sensory inputs or high-capacity neural networks, which limits their deployment on resource-constrained onboard platforms and increases training complexity. Second, sparse LiDAR observations often fail to explicitly highlight high-risk regions in cluttered environments, making policy training unstable and reducing generalization ability. Finally, most RL frameworks focus on maximizing expected rewards without explicitly modeling safety constraints, which may will lead to risky behaviors during training and deployment. Therefore, how to design a lightweight perception representation that can effectively utilize sparse LiDAR observations while ensuring safe and robust navigation remains an open problem for UAV autonomous flight in complex environments.

To address the above issues, We proposes a safety-constrained perception-control integrated framework for UAV autonomous navigation. The main contributions of our work are summarized as follows:
\begin{enumerate}
    \item We proposes a safety-constrained UAV navigation method to address instability and safety issues under sparse perception in environments with varying obstacle densities.
    \item The approach integrates a lightweight perception network, a unified state representation, and a hierarchical control framework, combined with a Lagrangian-based safe PPO algorithm and a curriculum learning strategy to enable efficient and safe decision-making.
    \item Experiments in environments with varying obstacle densities and flight speeds are conducted to demonstrate the improved success rate, enhanced safety, and higher efficiency of the proposed method compared with existing reinforcement learning baselines.
\end{enumerate}

\section{RELATED WORK}

Deep Reinforcement Learning (DRL) has been widely applied to autonomous UAV navigation and obstacle avoidance due to its ability to learn reactive control policies in complex environments. PPO-based methods exhibit stable training in continuous control tasks and have been used for UAV maneuvering and path planning\cite{wang2025research}. Hybrid frameworks combining model-based control and RL further improve adaptability in dynamic obstacle scenarios\cite{skarka2024hybrid}, while end-to-end RL enables policy learning in multi-UAV or complex urban environments without explicit global mapping\cite{pan2026action}. However, most existing approaches formulate the problem as a standard MDP and handle safety objectives implicitly via reward shaping. In dense obstacle environments, this often leads to unstable training and safety violations. Moreover, ensuring onboard real-time deployment and computational efficiency under high-dimensional perceptual inputs remains underexplored.

To explicitly incorporate safety constraints, UAV navigation is increasingly formulated as a CMDP\cite{kushwaha2025survey}. Lagrangian-relaxed PPO introduces dual variables to balance task reward and constraint cost dynamically\cite{achiam2017constrained}, achieving more stable constraint satisfaction than conventional penalty-based methods. Recent work improves stability via enhanced dual update schemes and safety value function estimation\cite{xu2026tcrl}, and further integrates control-theoretic tools such as control barrier functions for stronger safety guarantees\cite{ahmad2025hierarchical}. Nevertheless, in real UAV platforms, dual update sensitivity, approximation errors in safety value functions, and maintaining a stable trade-off between performance and safety in complex obstacle environments remain key challenges for Safe RL deployment.

LiDAR provides direct geometric information and is crucial for UAV navigation and obstacle avoidance. Traditional methods often rely on real-time lidar odometry and mapping frameworks, e.g., LOAM\cite{zhang2014loam} and Voxblox\cite{oleynikova2017voxblox}. With the rise of learning-based approaches, LiDAR observations are also used for end-to-end policy learning, though most studies assume high-line 3D LiDAR or high-resolution depth input. In contrast, lightweight UAV typically use sparse multi-line LiDAR, whose limited vertical resolution poses challenges for modeling obstacle continuity and identifying navigable gaps. Moreover, raw distance measurements do not explicitly encode collision risk monotonicity, potentially affecting RL training stability. Efficient perception modeling for sparse LiDAR thus remains an open research problem.

\begin{figure*}
    \centering
    \includegraphics[width=1.0\linewidth]{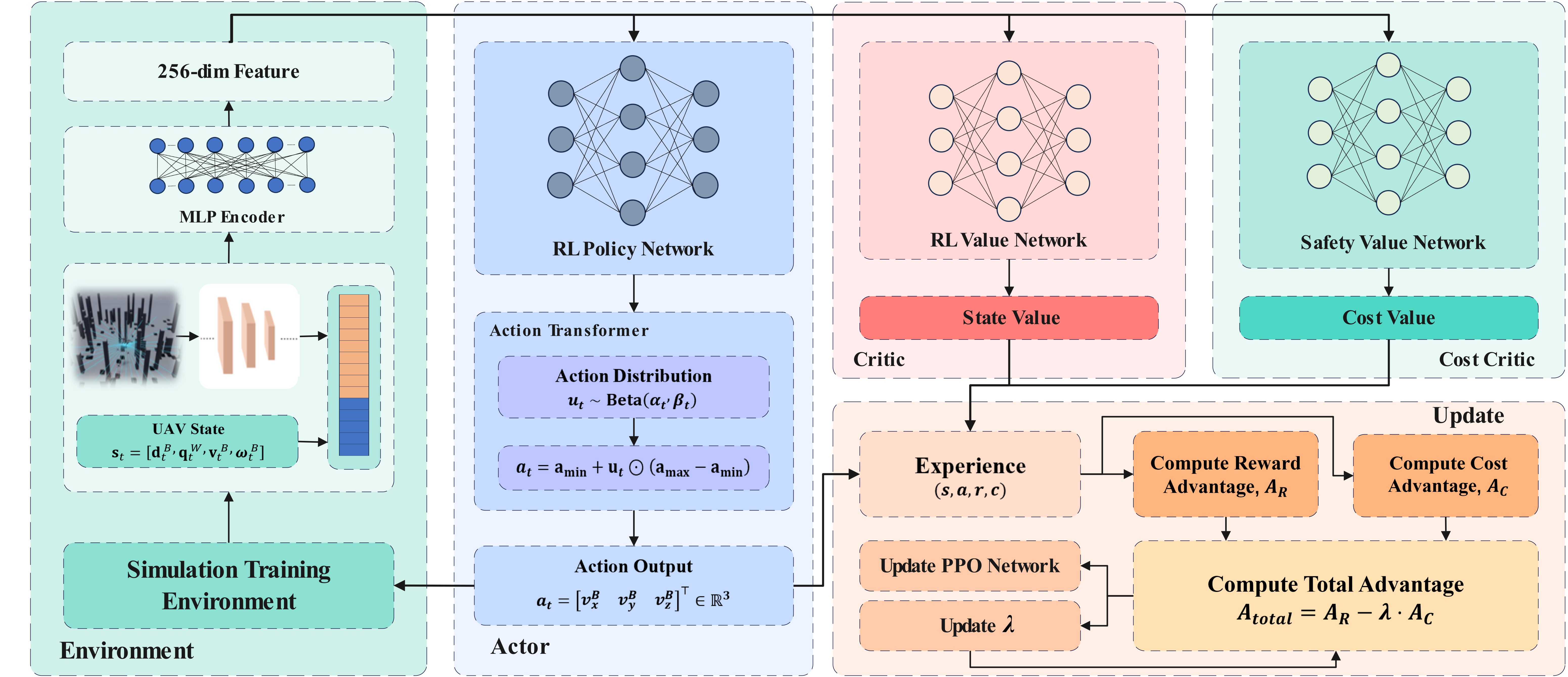}
    \caption{Overview of the proposed safety-aware PPO framework. Sparse LiDAR observations and UAV states are encoded into a shared latent representation that feeds the Actor, Critic, and Cost Critic. The Actor outputs bounded velocity commands via a Beta policy, while the Critic and Cost Critic estimate reward and safety cost, respectively. Policy updates are performed using a Lagrangian-based objective that balances task performance and safety.}
    \label{fig:placeholder}
\end{figure*}

\section{METHOD}

We formulate the UAV obstacle avoidance navigation problem as a CMDP. and propose a unified framework that integrates lightweight sparse LiDAR perception, safe reinforcement learning, and curriculum learning. The proposed method is built upon PPO, enabling end-to-end collision avoidance policy learning under explicit safety constraints, while leveraging efficient perception representations and staged training to improve stability and generalization in complex environments.

\subsection{Convolutional Feature Extraction from LiDAR Data}

\begin{figure}[H]
    \centering
    \includegraphics[width=1.0\linewidth]{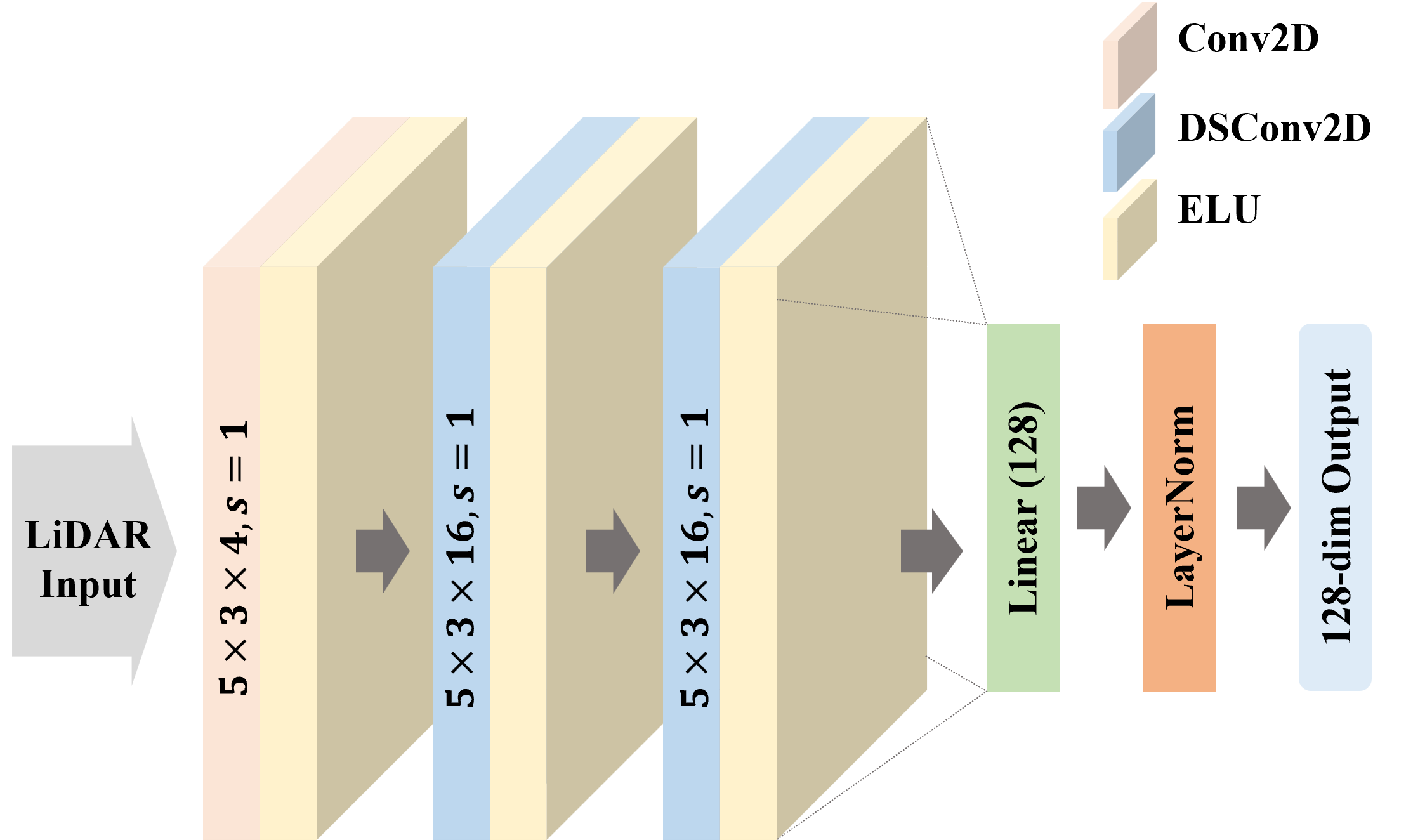}
    \caption{Architecture of the LiDAR Feature Extraction Network,where Conv and DSConv denote standard and depthwise-separable convolutions, respectively.}
    \label{fig:CNN}
\end{figure}

We propose a lightweight convolutional encoder to efficiently extract geometric features related to obstacle avoidance from LiDAR observations.The encoder learns representations from the input observation $X$ and produces compact feature embeddings for the policy network.The overall network is illustrated in Fig.~\ref{fig:CNN}.

The last two layers of the encoder adopt depthwise separable convolutions, decomposing standard convolutions into channel-wise spatial filtering and pointwise feature fusion operations \cite{howard2017mobilenets}. This design substantially reduces parameter count and computational complexity while preserving the capacity to model local geometric structures, making it well suited for resource-constrained onboard platforms. For sparse LiDAR tensors with limited spatial resolution, the architecture effectively captures horizontally continuous obstacle boundaries as well as height variations reflected by vertically distributed beams. Such a structure supports both large-scale simulation training and real-time deployment in practical UAV systems.

\begin{figure}
    \centering
    \includegraphics[width=1.0\linewidth]{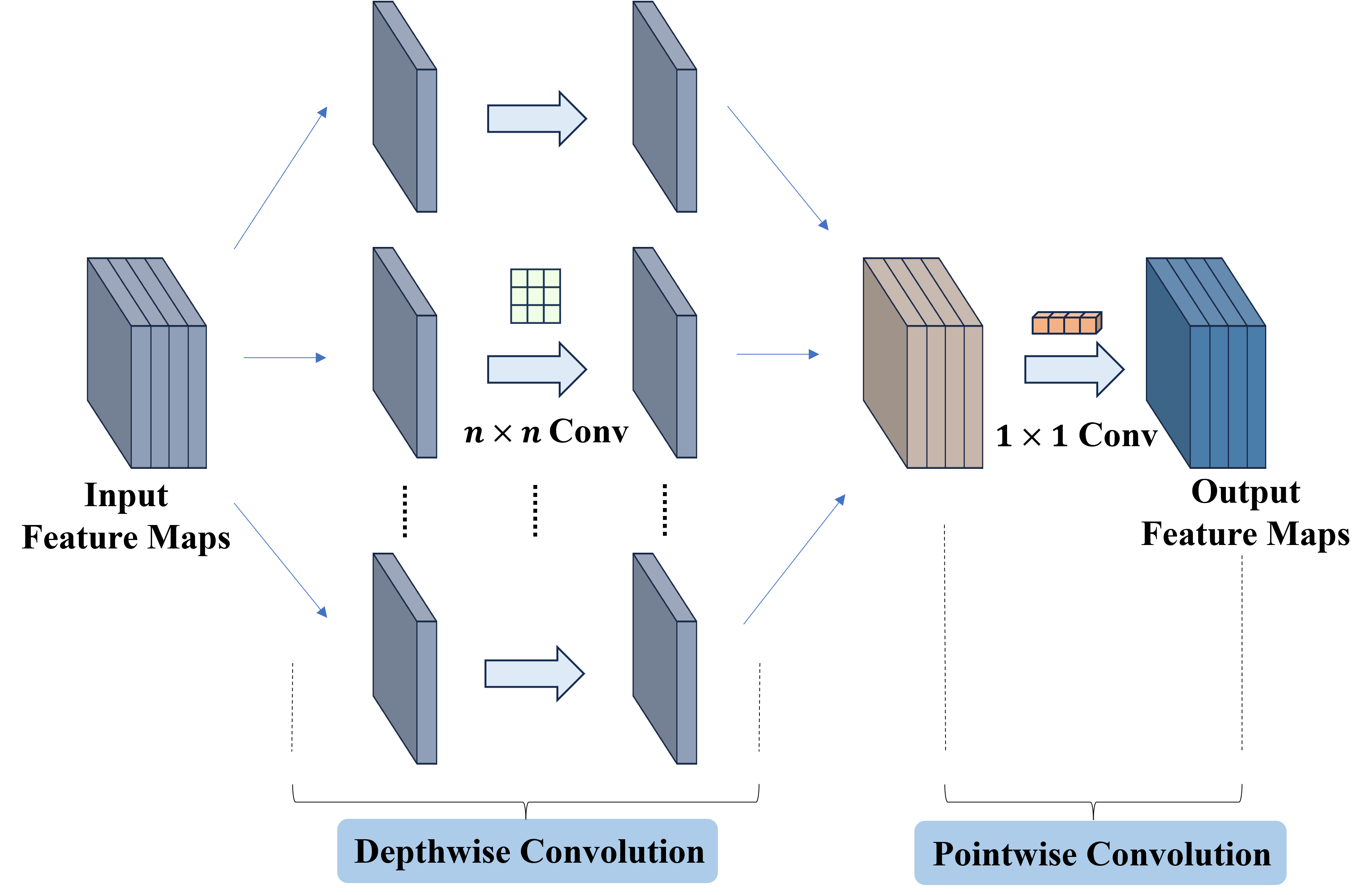}
    \caption{Schematic of Depthwise-Separable Convolution}
    \label{fig:DSC}
\end{figure}

The convolutional layers employ asymmetric kernels to enhance the modeling of horizontal angular resolution while explicitly accounting for the coupling between a limited number of vertical beams. Progressive downsampling is applied across layers to enlarge the receptive field and suppress redundant information. Finally, the convolutional features are mapped into a 128-dimensional embedding vector, followed by Layer Normalization to mitigate distribution shifts caused by variations in LiDAR return statistics across different environments.

\subsection{State Space Definition}

In our study, the observation space is designed to provide the agent with comprehensive information about its own motion state, the relative target pose, and the surrounding obstacle distribution, thereby supporting robust autonomous navigation in complex environments. The complete observation at time $t$ can be expressed as
\begin{equation}
o_t = (s_t, X_t)
\end{equation}
where $s_t$ denotes the proprioceptive state vector, and $X_t$ represents the environment perception information obtained and processed from external LiDAR sensors.

\subsubsection{Proprioceptive State}

To improve the policy's robustness to global heading variations and avoid excessive reliance on the world coordinate system during learning, key motion state variables are expressed in the drone's body frame ($B$). The proprioceptive state vector is defined as
\begin{equation}
s_t = \big[ d_t^{B}, q_t^{W}, v_t^{B}, \omega_t^{B} \big]
\end{equation}
where $d_t^{B}$ denotes the normalized direction vector pointing toward the goal, $q_t^{W}$ represents the attitude quaternion in the world coordinate frame, and $v_t^{B}$ and $\omega_t^{B}$ denote the linear velocity and angular velocity in the body frame, respectively. This design preserves essential dynamic information while effectively reducing the dependence of the state representation on absolute pose, thereby facilitating the learning of motion patterns with rotational invariance.

\subsubsection{Exteroceptive Perception}

External environmental perception is provided by an onboard LiDAR sensor. 
At each time step $t$, the observation is represented as a 2D tensor
\begin{equation}
X_t \in \mathbb{R}^{1 \times 36 \times 4}
\end{equation}
where the horizontal direction is discretized into 36 angular bins, 
and the vertical dimension corresponds to four laser beams with different elevation angles.

A proximity-based representation is used to convert raw LiDAR distances into features that better reflect collision risk by emphasizing nearby obstacles and reducing the influence of distant ones. This improves feature contrast and helps the convolutional network capture local geometric structures more effectively. The resulting proximity values are normalized to $[0,1]$ and arranged as an angle-elevation grid, which is then fused with the ego-state for policy learning.

\subsubsection{Shared Feature Encoder}

To effectively fuse geometric perception information with system dynamics, we employ a shared feature encoder $f_{\mathrm{enc}}(\cdot)$ to learn a unified representation of the observations. The LiDAR observation $X_t$ is first processed by the convolutional encoder described earlier to extract geometric features. The ego-state $s_t$ is linearly embedded, concatenated with the perception features, and then mapped by a multilayer perceptron into a latent representation
\begin{equation}
h_t = f_{\mathrm{enc}}(s_t, X_t)
\end{equation}
This latent representation is used as a shared input for the Actor, Critic, and Safety Value Network. Such a design reduces parameter redundancy, stabilizes the training process, and improves the generalization capability of the learned policy.

\subsection{Reward Function Design}

Efficient, smooth, and safe autonomous navigation in complex forest environments is encouraged by formulating the learning objective as the maximization of the discounted cumulative return.The overall reward function is defined as
\begin{equation}
r_t = r_{\text{vel}} + r_{\text{safety}} 
+ \lambda_h r_{\text{height}} 
+ \lambda_s r_{\text{smooth}} 
+ r_{\text{event}}
\end{equation}
where each term corresponds to task progress, safety constraints, flight height regulation, action smoothness, and sparse event rewards, respectively. The coefficients $\lambda_h$ and $\lambda_s$ control the relative importance of the corresponding terms.

\subsubsection{Velocity Tracking Reward}
We introduce a velocity projection-based reward to encourage the UAV to move toward the goal direction:
\begin{equation}
r_t^{\text{track}} =
\operatorname{clip}\left(
\mathbf{v}_t \cdot \mathbf{d}_t,
-v_{\max},
v_{\max}
\right)
\end{equation}
where $\mathbf{v}_t$ denotes the current linear velocity, 
$\mathbf{d}_t$ is the unit direction vector toward the goal, and 
$v_{\max}$ is a truncation threshold to prevent the agent from obtaining excessive rewards through overly high speeds, which may increase collision risk.

\subsubsection{LiDAR-based Safety Reward}
We construct a logarithmic obstacle potential reward based on LiDAR proximity to encourage the agent to actively maintain a safe distance from obstacles:
\begin{equation}
r_{\text{safety}} =
\mathbb{E}_{i \in \mathcal{S}}
\left[
\log
\left(
\operatorname{clip}
\left(
d_{\text{scan},i},
\epsilon,
R
\right)
\right)
\right]
\end{equation}
where $\mathcal{S}$ denotes the set of LiDAR beams, $R$ is the maximum sensing range, 
$d_{\text{scan},i}$ is the measured distance of the $i$-th beam, and $\epsilon$ is a small constant for numerical stability. Intuitively, this term encourages the agent to maximize the average residual safe distance from surrounding obstacles.

\subsubsection{Height Constraint Reward}
To prevent the UAV from flying too high or touching the ground, a quadratic penalty centered on a safe altitude interval is introduced:
\begin{equation}
r_t^{\text{height}} =
\begin{cases}
-(z_t - z_{\max})^2, & z_t > z_{\max}, \\
-(z_t - z_{\min})^2, & z_t < z_{\min}, \\
0, & \text{otherwise}.
\end{cases}
\end{equation}
where $z_t$ denotes the current flight altitude.

\subsubsection{Action Smoothness Reward}
We introduce a penalty on velocity variation to reduce aggressive maneuvers and improve energy efficiency:
\begin{equation}
r_t^{\text{smooth}} =
- \left\|
\mathbf{v}_t - \mathbf{v}_{t-1}
\right\|^2
\end{equation}

\subsubsection{Sparse Event Reward}
At the end of each episode, discrete rewards or penalties are applied based on the final state:
\begin{itemize}
\item \textbf{Goal reached:} A reward of $+100$ is given when the UAV reaches the goal within a distance of $0.5\,\text{m}$.
\item \textbf{Collision:} A penalty of $-50$ is applied when a collision with obstacles is detected.
\item \textbf{Abnormal state:} A penalty of $-20$ is assigned when the UAV exits the safety boundary (e.g., excessively high or low altitude).
\end{itemize}

\subsection{Safe Reinforcement Learning Algorithm Design}

In our work, we adopt a Lagrangian relaxation-based safe Proximal Policy Optimization (Safe PPO) algorithm as the decision-making framework. Built upon the standard PPO algorithm, explicit safety constraints are introduced, and a dual optimization mechanism is employed to dynamically balance task performance and safety. This framework is designed to enable safe and autonomous UAV navigation in complex obstacle-rich environments.

\subsubsection{Actor Network and Action Modeling}

We adopt a hierarchical control architecture, where the RL policy generates high-level velocity commands and a geometric controller tracks the desired motion and stabilizes the UAV dynamics.

The action space is defined as the desired body-frame velocity
\begin{equation}
\mathbf{a}_t =
\begin{bmatrix}
v_x^B & v_y^B & v_z^B
\end{bmatrix}^{\top}
\in \mathbb{R}^3
\end{equation}
which provides rotational invariance across different global headings.

Since the action space is continuous and bounded, the actor models the policy using a Beta distribution to avoid the distortion caused by Gaussian action clipping.
Given the latent feature $h_t$, the actor predicts the distribution parameters
\begin{equation}
\alpha_t = \operatorname{Softplus}(z_{\alpha}) + \epsilon,
\quad
\beta_t = \operatorname{Softplus}(z_{\beta}) + \epsilon.
\end{equation}

The action is sampled from
\begin{equation}
\mathbf{u}_t \sim \operatorname{Beta}(\alpha_t, \beta_t)
\end{equation}
and mapped to the valid velocity range
\begin{equation}
\mathbf{a}_t =
\mathbf{a}_{\min}
+
\mathbf{u}_t
\odot
(\mathbf{a}_{\max} - \mathbf{a}_{\min})
\end{equation}
where $\odot$ denotes element-wise multiplication.

The resulting velocity command is transformed to the world frame and tracked using the geometric controller proposed by Lee et al\cite{lee2010control}.

\subsubsection{Critic Network Structure}

The Critic network is used to estimate the state value function $V(s)$ and provides a baseline for Generalized Advantage Estimation (GAE). Based on the shared latent representation $h_t$, the Critic outputs a scalar state value through a single linear layer. To improve numerical stability during the early stage of training, orthogonal initialization is applied to the output layer of the Critic.

\subsubsection{Safety Value Network (Cost Network)}

To strictly limit risky behaviors of the UAV while maximizing task rewards, we introduce a safety cost function within the CMDP framework
\begin{equation}
C_{\text{total}} =
I_{\text{collision}} +
w \cdot C_{\text{margin}}
\end{equation}
where $I_{\text{collision}}$ represents a hard constraint signal indicating collision or severe boundary violation.

The safety margin term is defined as
\begin{equation}
C_{\text{margin}} =
\frac{
\operatorname{clamp}
\left(
d_{\text{safe}} - d_{\min},\, 0
\right)
}{d_{\text{safe}}}
\end{equation}
where $d_{\min}$ denotes the minimum obstacle distance detected by LiDAR and $d_{\text{safe}}$ is a predefined safety threshold. When $d_{\min} < d_{\text{safe}}$, a linear penalty is applied, providing a differentiable gradient signal before collision occurs.

Formally, our objective is to find an optimal policy $\pi$ that maximizes the expected cumulative reward while satisfying an expected cost constraint:
\begin{equation}
\begin{aligned}
\max_{\pi} \quad
J_R(\pi)
&=
\mathbb{E}_{\tau \sim \pi}
\left[
\sum_{t=0}^{T}
\gamma^t r_t
\right],
\\
\text{s.t.} \quad
J_C(\pi)
&=
\mathbb{E}_{\tau \sim \pi}
\left[
\frac{1}{T}
\sum_{t=0}^{T}
C_t
\right]
\le d_{\text{limit}}
\end{aligned}
\end{equation}
where
\begin{itemize}
\item $J_R(\pi)$ denotes the expected cumulative reward,
\item $J_C(\pi)$ represents the expected average cost per step,
\item $C_t$ is the safety cost defined above,
\item $d_{\text{limit}}$ is the predefined safety tolerance threshold.
\end{itemize}

Using Lagrangian relaxation, we introduce a multiplier $\lambda$ and convert the constrained problem into the following dual optimization problem:
\begin{equation}
\min_{\lambda}
\max_{\pi}
\mathcal{L}(\pi,\lambda)
=
J_R(\pi)
-
\lambda
\left(
J_C(\pi) - d_{\text{limit}}
\right)
\end{equation}

To compute gradients of the Lagrangian objective under the PPO framework \cite{schulman2017proximal}, we introduce a safety value network (Cost Critic), denoted as $V_{\phi}^{C}(s_t)$, which operates in parallel with the reward value network $V_{\theta}^{\pi}(s_t)$. The Cost Critic estimates the expected cumulative cost from the current state. Its training objective is
\begin{equation}
\mathcal{L}_{\text{Cost}}(\phi)
=
\mathbb{E}_t
\left[
\left(
V_{\phi}^{C}(s_t) -
\hat{R}_t^{C}
\right)^2
\right]
\end{equation}
where $\hat{R}_t^{C}$ denotes the cost return computed from sampled trajectories.

Based on these two value networks, we compute the reward advantage $A_R^{\pi}(s,a)$ and the cost advantage $A_C^{\pi}(s,a)$. To reduce variance, both advantages are estimated using Generalized Advantage Estimation (GAE).

\subsubsection{Safe PPO Objective}

Under the PPO framework, we define a joint advantage function:
\begin{equation}
A_{\text{total}}(s,a)
=
\hat{A}_R^{\text{GAE}}(s,a)
-
\lambda
\hat{A}_C^{\text{GAE}}(s,a)
\end{equation}

The clipped policy objective is then formulated as
\begin{equation}
\begin{aligned}
\mathcal{L}_{\text{policy}}(\theta)
&=
\mathbb{E}_t
\Big[
\min \Big(
\rho_t(\theta) A_{\text{total}}, \\
&\quad
\operatorname{clip}
\big(
\rho_t(\theta),
1-\epsilon,
1+\epsilon
\big)
A_{\text{total}}
\Big)
\Big]
\end{aligned}
\end{equation}
where
\begin{equation}
\rho_t(\theta)
=
\frac{
\pi_{\theta}(a_t | s_t)
}{
\pi_{\theta_{\text{old}}}(a_t | s_t)
}
\end{equation}

This formulation suppresses high-risk actions even if they yield high reward advantages, due to the cost penalty.

\subsubsection{Adaptive Update of the Lagrange Multiplier}

The Lagrange multiplier $\lambda$ is updated using dual gradient ascent:
\begin{equation}
\lambda_{k+1}
=
\operatorname{ReLU}
\left(
\lambda_k
+
\eta_{\lambda}
\left(
\bar{C}_k
-
d_{\text{limit}}
\right)
\right)
\end{equation}
where $\bar{C}_k$ denotes the average cost of the current iteration. This mechanism increases the constraint strength when safety violations occur frequently and relaxes it when the policy becomes safer.

Additionally, a linear learning rate decay is applied to both the actor and critic during training to improve convergence stability.

\subsection{Curriculum Learning via Feature Reuse and Policy Reset}

We adopt a feature-transfer-based curriculum learning strategy to mitigate training instability caused by increasing obstacle density\cite{bengio2009curriculum}. The model is first pretrained in a low-density environment to learn stable geometric representations and fundamental dynamics control capabilities. It is then transferred to a high-density environment with the pretrained weights loaded.

During the transfer stage, the parameters of the shared feature encoder are preserved to reuse the learned environmental representation capability, while the output layer of the Actor network is orthogonally reinitialized. This design prevents the previous policy from overfitting or converging to suboptimal solutions in the new environment and restores the policy to a high-entropy exploration regime. Supported by the pretrained feature space, the agent can converge more sample-efficiently to a safe policy adapted to high-density scenarios.

\section{Experiment}

\subsection{Simulation Environment and Task Setup}

We evaluate the effectiveness of the proposed method in complex unstructured environments by constructing an autonomous UAV navigation task in a high-fidelity 3D forest-like environment using the NVIDIA Isaac Sim simulation platform together with the OmniDrones framework\cite{chen2025matters}.

\begin{figure}[h]
    \centering
    \includegraphics[width=\columnwidth]{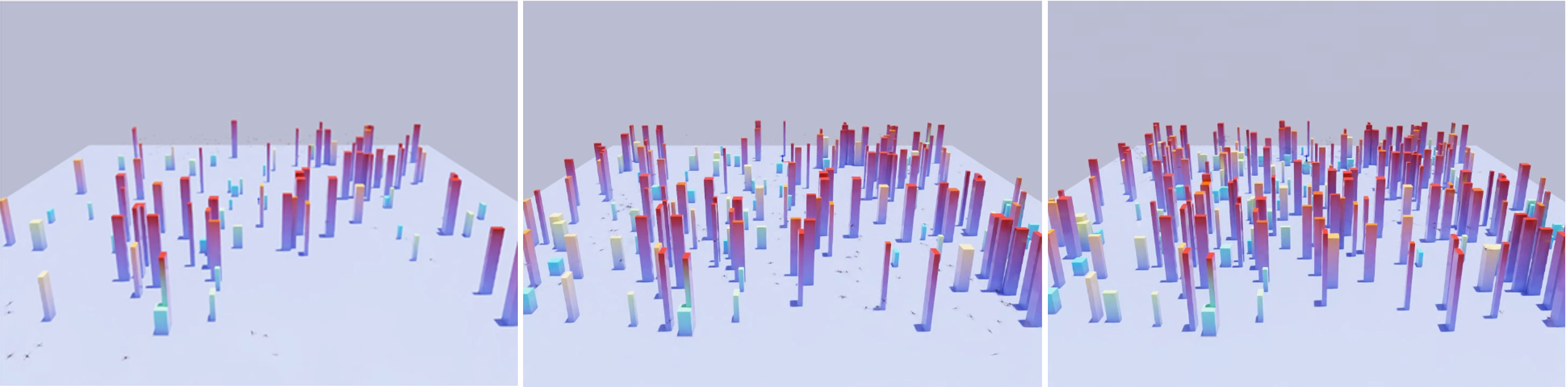}
    \caption{Simulation environments with varying obstacle densities (100, 200, and 300 obstacles from left to right).}
    \label{fig:environment}
\end{figure}

The simulation environment is a flat $40\,\mathrm{m} \times 40\,\mathrm{m}$ area, as shown in Fig.~\ref{fig:environment}. Cylindrical obstacles with numbers of 100, 200, and 300 are randomly placed on the ground to represent environments from sparse to highly cluttered. Obstacle geometries are randomized to increase scene diversity: widths are uniformly sampled from $0.4$-$1.0\,\mathrm{m}$, and heights follow a piecewise distribution over $[1.0,2.0)$, $[2.0,4.0)$, and $[4.0,5.0]\,\mathrm{m}$, producing multi-scale occlusions and altitude-dependent collision risks.

The UAV is equipped with a lightweight sparse multi-beam LiDAR sensor with a maximum sensing range of $4.0\,\mathrm{m}$. The observations are organized as a distance tensor of size $1 \times 36 \times 4$, covering the full horizontal field of view and a limited vertical field of view. The agent starts from a random initial position and autonomously navigates toward a target location while maintaining a specified flight altitude and avoiding collisions with obstacles, the ground, and the environment boundaries.

This task evaluates both obstacle-avoidance responsiveness and goal-directed navigation stability under extremely sparse perception conditions. To ensure reproducibility, all environment generation processes are conducted with a fixed random seed, and the environment 
scale, obstacle distribution rules, and sensor configuration remain consistent across all experiments.

\subsection{Evaluation Metrics}

Autonomous navigation performance in complex unstructured environments is quantitatively evaluated from two complementary perspectives: task completion capability and flight safety. The evaluation metrics are defined as follows:

\begin{enumerate}
\item \textbf{Success Rate (SR)}  
The success rate is defined as the proportion of test episodes in which the UAV successfully completes the navigation task without any collision. This metric reflects the overall task completion capability of the policy.

\item \textbf{Collision Rate (CR)}  
The collision rate represents the proportion of test episodes that terminate prematurely due to collisions with obstacles, the ground, or the environment boundaries. It is used to evaluate the safety performance of the policy in complex environments.

\item \textbf{Average Return (AR)}  
The average return is defined as the mean cumulative reward obtained per episode during the evaluation phase. Provides a comprehensive measure of the performance of the policy in terms of task efficiency and compliance with safety constraints.

\item \textbf{Average Episode Length (AEL)}  
This metric denotes the average number of steps (or the equivalent flight distance) per episode during evaluation, reflecting the stability and efficiency of the navigation policy.
\end{enumerate}

\begin{table*}
\centering
\caption{Module impact on navigation performance under different obstacle densities.}
\label{tab:ablation}
\setlength{\tabcolsep}{12pt}      % 列间距变大
\begin{tabular}{lccc|ccc}
\toprule
Method & DSC & Safe RL & Curriculum & SR (100) & SR (200) & SR (300) \\
\midrule
Vanilla PPO & $\times$ & $\times$ & $\times$ & 0.83594 & 0.69531 & 0.60938 \\
PPO + DSC & $\checkmark$ & $\times$ & $\times$ & 0.86719 & 0.70313 & 0.66406 \\
PPO + DSC + Safe RL & $\checkmark$ & $\checkmark$ & $\times$ & 0.96875 & 0.90625 & 0.86719 \\
Ours (Full) & $\checkmark$ & $\checkmark$ & $\checkmark$ & \textbf{0.96875} & \textbf{0.95313} & \textbf{0.94531} \\
\bottomrule
\end{tabular}
\end{table*}

\subsection{Module Contribution under Increasing Complexity}
We conduct module-wise evaluations in three forest scenarios containing 100, 200, and 300 obstacles to systematically assess their contributions under varying environmental complexity.All methods share the same training configuration and evaluation metrics and are progressively compared, including Vanilla PPO, PPO+DSC, PPO+DSC+Safe RL, and the full model with curriculum learning (Ours). The curriculum learning strategy follows a staged transfer scheme, where training in higher-density environments is initialized from models that have converged in lower-density environments, thereby gradually increasing task difficulty.

The quantitative results are presented in Table~\ref{tab:ablation}. Vanilla PPO achieves a success rate of 0.83594 in the 100 obstacles scenario, but its performance decreases to 0.69531 and 0.60938 in the 200 and 300 obstacles environments, indicating limited capability in modeling complex obstacle structures under sparse LiDAR observations. As obstacle density increases, navigation becomes more challenging due to fragmented observations, occlusions, and reduced free-space continuity, which further exposes the limitations of the baseline policy in cluttered environments. After introducing depthwise separable convolutions, performance improves across all scenarios, with more noticeable gains in the 300 obstacles setting. This result suggests that the proposed feature extraction module can better capture local geometric continuity and structural information from sparse LiDAR data while maintaining computational efficiency. Incorporating Safe RL further increases the success rates to 0.96875, 0.90625, and 0.86719, respectively. The improvements are particularly evident in medium- and high-density environments, highlighting the stabilizing effect of explicit safety constraints and their ability to guide the policy toward safer navigation behaviors during training.

Finally, integrating curriculum learning leads to additional gains in the 200 and 300 obstacles scenarios, achieving success rates of 0.95313 and 0.94531, effectively mitigating performance degradation as environmental complexity increases. By gradually increasing the difficulty of training environments, the agent is able to progressively adapt to more complex obstacle distributions and learn more robust navigation strategies. Notably, the improvements become more significant as the obstacle density increases, particularly in the 300 obstacles scenario, indicating that the proposed framework scales well to highly cluttered environments. Overall, these results demonstrate that the proposed components play complementary roles: the perception module enhances representation capability under sparse observations, the safety mechanism stabilizes policy optimization, and curriculum learning improves adaptability and generalization. Together, they jointly improve navigation robustness and reliability in complex unstructured environments.

\subsection{Robust Navigation under Increasing Obstacle Density}

To evaluate performance in complex unstructured environments, we compare the proposed full model (Ours) with representative reinforcement learning baselines, including PPO, off-policy methods SAC\cite{haarnoja2018soft} and TD3\cite{fujimoto2018addressing}, and recurrent variants PPO+GRU\cite{zhang2024autonomous} and PPO+LSTM\cite{ghojogh2023recurrent}. We further investigate recurrent extensions of our framework (Ours+GRU and Ours+LSTM). All methods are evaluated in environments containing 100, 200, and 300 obstacles. As shown in Table~\ref{tab:success_rate}, most baseline methods exhibit noticeable performance degradation as obstacle density increases, whereas our method consistently achieves the highest success rates, demonstrating stronger robustness and stability in cluttered scenarios. Although incorporating GRU or LSTM introduces only marginal performance changes, it significantly increases computational cost. Ours+GRU and Ours+LSTM require approximately 2.4× and 2.5× longer training time than the proposed model, while SAC and TD3 require about 2.7× more time to reach stable performance.
\begin{table}[H]
\centering
\caption{ Comparison with baseline RL methods under different
obstacle densities.}
\label{tab:success_rate}

\renewcommand{\arraystretch}{1.2} % 行距变大
\setlength{\tabcolsep}{12pt}      % 列间距变大

\begin{tabular}{lccc}
\toprule
\textbf{Method} & \textbf{SR (100)} & \textbf{SR (200)} & \textbf{SR (300)} \\
\midrule
PPO        & 0.83594 & 0.69531 & 0.60938 \\
SAC        & 0.86719 & 0.73438 & 0.66406 \\
TD3        & 0.92969 & 0.89063 & 0.79688 \\
PPO+GRU    & 0.85938 & 0.71875 & 0.61719 \\
PPO+LSTM   & 0.87500 & 0.75781 & 0.73438 \\
Ours       & 0.96875 & 0.95313 & 0.94531 \\
Ours+GRU   & 0.97656 & 0.96875 & 0.91406 \\
Ours+LSTM  & 0.95313 & 0.92969 & 0.88281 \\
\bottomrule
\end{tabular}
\end{table}

In addition to training efficiency, model complexity is further analyzed. As reported in Table~\ref{tab:params}, the proposed model contains only 143k parameters in the main network, substantially fewer than SAC, TD3, and recurrent PPO variants, whose parameter counts typically exceed one million due to additional critic and target networks. This lightweight design reduces computational overhead and accelerates convergence, making the framework more suitable for resource-constrained onboard deployment. Considering performance, efficiency, and model complexity together, we adopt the non-recurrent version as the final model, indicating that the proposed sparse-perception-aware representation and safety-constrained learning mechanism are sufficient for robust navigation under sparse LiDAR observations.
\begin{table}[H]
\centering
\caption{Network Parameter Comparison of Different Methods}
\label{tab:params}
\renewcommand{\arraystretch}{1.2} % 行距变大
\begin{tabular}{lcc}
\toprule
\textbf{Method} & \textbf{Main Network Params} & \textbf{Including Target Networks} \\
\midrule
Ours        & 143,208   & -- \\
SAC         & 694,314   & 1,245,934 \\
TD3         & 693,543   & 1,245,163 \\
PPO+GRU     & 1,073,567 & -- \\
PPO+LSTM    & 1,337,759 & -- \\
Ours+GRU    & 1,609,708 & -- \\
Ours+LSTM   & 2,005,996 & -- \\
\bottomrule
\end{tabular}
\end{table}

\begin{figure}
    \centering
    \includegraphics[width=1.0\linewidth]{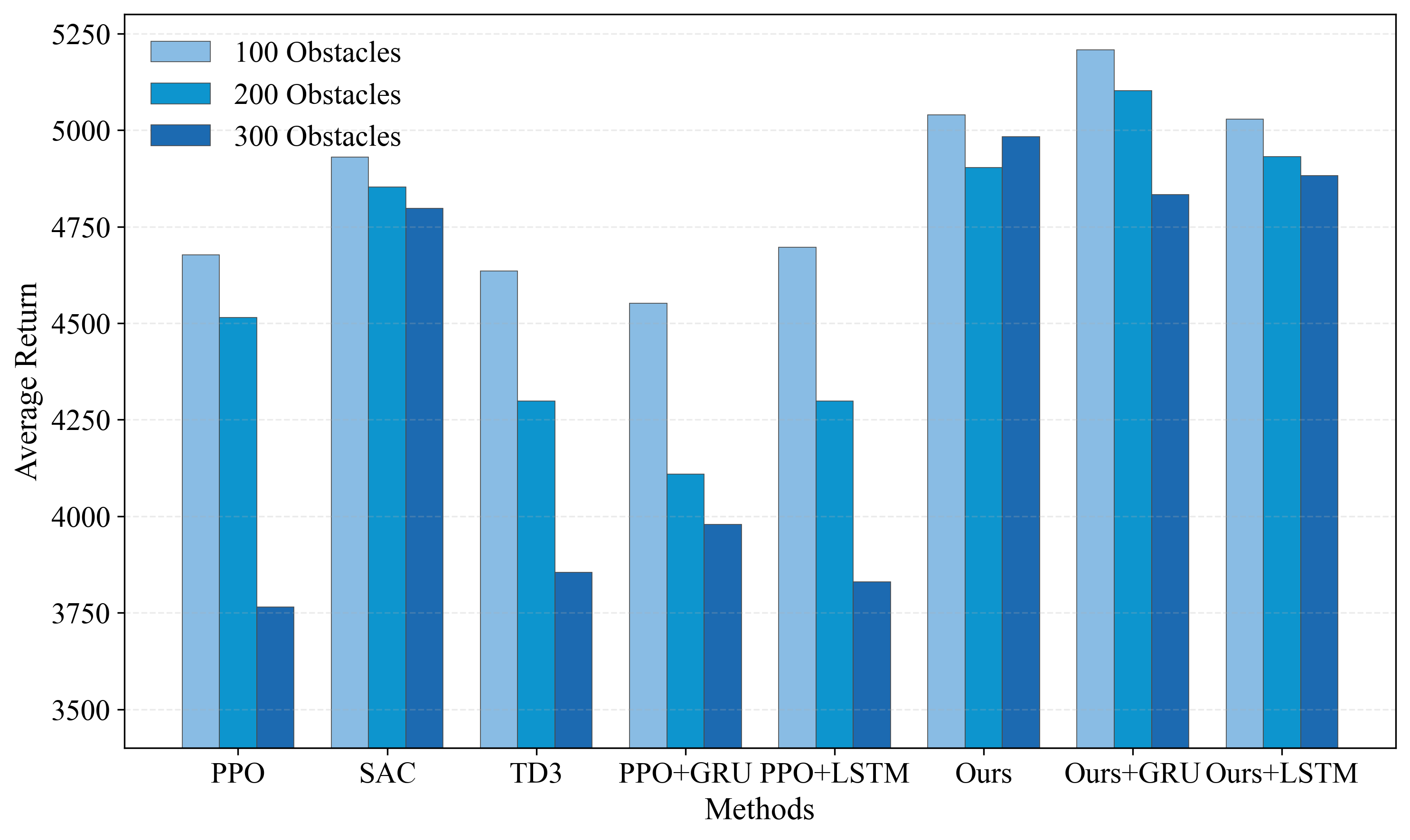}
    \caption{Comparison with baseline RL methods under different obstacle densities.}
    \label{fig:return}
\end{figure}

Fig.~\ref{fig:return} further reports the quantitative comparison in terms of average return for environments containing 100, 200, and 300 obstacles. 
Overall, our method achieves higher returns across all scenarios, indicating stronger robustness in cluttered environments. 
In contrast, TD3 and recurrent PPO variants exhibit noticeable performance degradation as the obstacle density increases, particularly in the 300 obstacles setting. 
These results demonstrate that the proposed framework maintains stable navigation performance even under significantly increased environmental complexity.

\subsection{High-Speed Flight under Constraints}
We evaluate robustness under varying dynamic and environmental conditions by analyzing the navigation success rate (SR) across maximum flight velocities of 2–11 m/s and obstacle densities of 100, 200, and 300, as illustrated in Fig.~6. Overall, the policy maintains high performance across a wide velocity range, with SR remaining around or above 0.94 in most cases. In particular, the best performance is observed near 6 m/s in the 100 obstacles environment, where the SR approaches 0.99, indicating a favorable balance between maneuverability and control stability. This also suggests that the learned policy can effectively exploit dynamic feasibility while maintaining reliable obstacle avoidance. As the velocity increases beyond this range, the SR gradually decreases, and the degradation becomes more pronounced in denser environments. For example, in the 300 obstacles scenario, the SR shows a noticeable drop when the velocity exceeds 8 m/s, reflecting the increased difficulty of collision avoidance under limited reaction time and perception uncertainty. This trend highlights the growing challenge of safe navigation when both environmental complexity and dynamic constraints increase simultaneously. Nevertheless, the proposed method still maintains competitive performance even at higher velocities, demonstrating strong robustness and stable generalization across different dynamic constraints.
\begin{figure}
    \centering
    \includegraphics[width=0.9\linewidth]{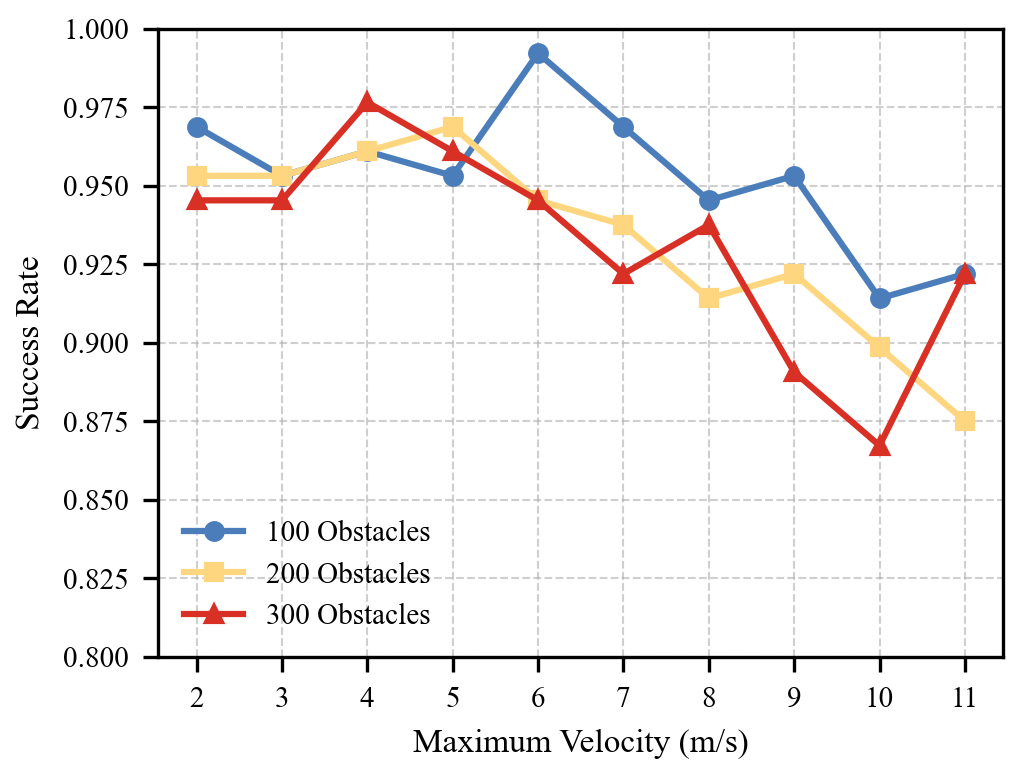}
    \caption{Navigation success rate under different maximum flight velocities.}
    \label{fig:velocity_analysis}
\end{figure}

\section{CONCLUSIONS}

We present a safe reinforcement learning framework for UAV navigation under sparse LiDAR perception. The method integrates a lightweight perception encoder, safety-constrained policy optimization, and curriculum learning to enhance training stability and navigation robustness. Experiments show consistent improvements over state-of-the-art reinforcement learning baselines across different obstacle densities and flight speeds. Future work will explore more diverse environments and improve robustness to sensing noise.

\addtolength{\textheight}{-12cm}   % This command serves to balance the column lengths
                                  % on the last page of the document manually. It shortens
                                  % the textheight of the last page by a suitable amount.
                                  % This command does not take effect until the next page
                                  % so it should come on the page before the last. Make
                                  % sure that you do not shorten the textheight too much.

%%%%%%%%%%%%%%%%%%%%%%%%%%%%%%%%%%%%%%%%%%%%%%%%%%%%%%%%%%%%%%%%%%%%%%%%%%%%%%%%

%%%%%%%%%%%%%%%%%%%%%%%%%%%%%%%%%%%%%%%%%%%%%%%%%%%%%%%%%%%%%%%%%%%%%%%%%%%%%%%%

%%%%%%%%%%%%%%%%%%%%%%%%%%%%%%%%%%%%%%%%%%%%%%%%%%%%%%%%%%%%%%%%%%%%%%%%%%%%%%%%

%%%%%%%%%%%%%%%%%%%%%%%%%%%%%%%%%%%%%%%%%%%%%%%%%%%%%%%%%%%%%%%%%%%%%%%%%%%%%%%%

\bibliographystyle{ieeetr}
\bibliography{name.bib}

\end{document}